\documentclass[sigconf,nonacm]{acmart}

\usepackage{algorithm}
\usepackage[noend]{algpseudocode}
\usepackage{multicol}
\usepackage{tikz}
\usepackage{subcaption}
\usepackage{xcolor}
\usepackage{siunitx}
\usepackage{graphicx}
\usepackage{cancel}
\usetikzlibrary{positioning,calc}

\AtBeginDocument{%
  }

\begin{document}

\title{Dimensional Peeking for Low-Variance Gradients in\\Zeroth-Order Discrete Optimization via Simulation}

\author{Philipp Andelfinger}
\author{Wentong Cai}
\email{{philipp.andelfinger, aswtcai}@ntu.edu.sg}
\affiliation{
  \institution{Nanyang Technological University}
  \city{Singapore}
  \country{Singapore}
}

\renewcommand{\shortauthors}{Andelfinger and Cai}

\begin{abstract}
Gradient-based optimization methods are commonly used to identify local optima in high-dimensional spaces.
When derivatives cannot be evaluated directly, stochastic estimators can provide approximate gradients.
However, these estimators' perturbation-based sampling of the objective function introduces variance that can lead to slow convergence.
In this paper, we present dimensional peeking, a variance reduction method for gradient estimation in discrete optimization via simulation.
By lifting the sampling granularity from scalar values to classes of values that follow the same control flow path, we increase the information gathered per simulation evaluation.
Our derivation from an established smoothed gradient estimator shows that the method does not introduce any bias.
We present an implementation via a custom numerical data type to transparently carry out dimensional peeking over C++ programs.
Variance reductions by factors of up to 7.9 are observed for three simulation-based optimization problems with high-dimensional input.
The optimization progress compared to three meta-heuristics shows that dimensional peeking increases the competitiveness of zeroth-order optimization for discrete and non-convex simulations.
\end{abstract}

%\begin{CCSXML}
%<ccs2012>
%<concept>
%<concept_id>10003752.10003809.10003716.10011136.10011797.10011801</concept_id>
%<concept_desc>Theory of computation~Randomized local search</concept_desc>
%<concept_significance>500</concept_significance>
%</concept>
%<concept>
%<concept_id>10002950.10003714.10003715.10003724</concept_id>
%<concept_desc>Mathematics of computing~Numerical differentiation</concept_desc>
%<concept_significance>500</concept_significance>
%</concept>
%</ccs2012>
%\end{CCSXML}
%
%\ccsdesc[500]{Theory of computation~Randomized local search}
%\ccsdesc[500]{Mathematics of computing~Numerical differentiation}
%
%\keywords{Simulation-based optimization, discrete optimization, gradient estimation, variance reduction}

\maketitle

\section{Introduction}

Simulation-based optimization (SBO) problems appear in many scientific contexts and application domains.
Their defining characteristic is that the objective function is a simulation, in contrast to the explicit analytical expression available in other forms of mathematical optimization~\cite{fu2015handbook}.
In many instances, additional properties of the objective function such as high input dimensionality, stochasticity, non-convexity, high cost of function evaluations, and the unavailability of derivatives make SBO particularly challenging.

Commonly, SBO problems are tackled in a black-box manner using response surface methods or meta-heuristics such as evolutionary algorithms~\cite{tekin2004simulation}.
In the past few years, there has been a renewed interest in solving SBO using gradient descent.
By various forms of smoothing, gradient estimates can be determined even for simulations involving discontinuities such as those introduced by conditional control flow~\cite{gong1987smoothed,arya2022automatic,chopra2023agenttorch,andelfinger2023towards,kreikemeyer2023smoothing}.
Beyond SBO, simulation gradients also enable a more natural integration of simulations into machine learning pipelines, e.g., for reinforcement learning~\cite{mora2021pods}.

A significant issue in gradient estimation over simulations is the high variance stemming from the diversity in state trajectories, which can limit the convergence speed.
This issue is particularly pressing in derivative-free, or \emph{zeroth-order}, optimization methods~\cite{larson2019derivative}, which often introduce additional stochasticity by random perturbations to the decision variables to estimate gradients of a smooth approximation of the original objective function.

\begin{figure}[t!]
  \centering
  \begin{subfigure}[b]{0.24\textwidth}
    \centering
    \hspace{-0.3cm}
    \includegraphics[width=0.76\textwidth]{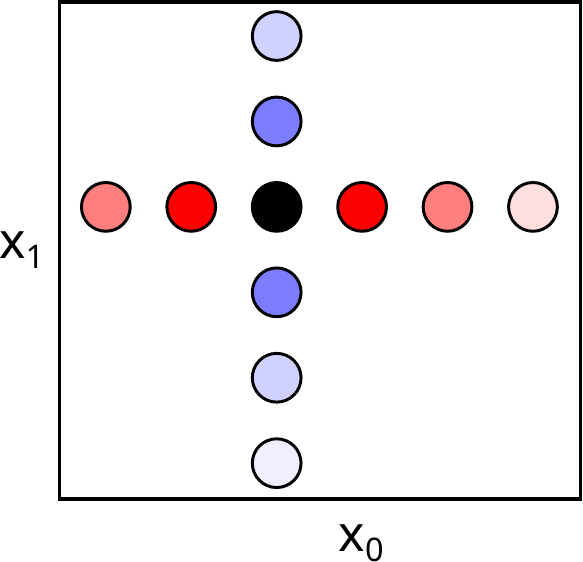}
    \vspace{-0.105cm}
    \caption{Augmented decision variables.}
  \end{subfigure}
  \hfill
  \begin{subfigure}[b]{0.22\textwidth}
    \centering
    \includegraphics[width=0.9\textwidth]{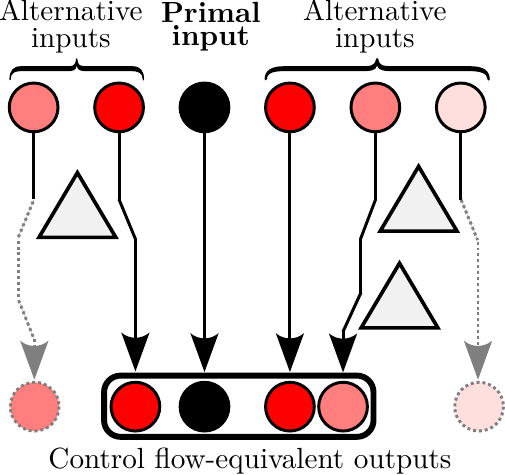}
    %\vspace{0.05cm}
    \caption{Dimensional peeking for $x_0$.}
  \end{subfigure}
  \caption{Dimensional peeking in a two-dimensional space. A primal perturbed input (black circle) is augmented by all alternative values of non-negligible probability per dimension. By identifying and grouping input values that follow the same path at branches (triangles), the sampling granularity is lifted from scalars to control flow-equivalent classes.}
  \label{fig:illustration}
  \vspace{-0.2cm}
\end{figure}

In the present paper, we propose a variance reduction method for zeroth-order optimization over discrete decision variables.
The main idea is to lift the simulation evaluation from the level of points in the input parameter space towards \emph{classes of points} that follow the same control flow path, thereby extracting significantly more information per evaluation.
Figure~\ref{fig:illustration} illustrates our method, which we refer to as \emph{dimensional peeking}, on a conceptual level.
The simulation output under the alternative perturbations is evaluated in a vectorized fashion alongside a ``primal'' perturbed evaluation, maintaining low overhead.
On each input dimension, all perturbations equivalent in control flow are considered jointly while exploiting a priori knowledge of the perturbation probabilities.
In effect, dimensional peeking eliminates the variance among per-dimension perturbations that do not differ in control flow.

Our main contributions are as follows:
\begin{itemize}
\item We present \textbf{dimensional peeking}, establish its unbiasedness with respect to an established smoothed gradient estimator~\cite{polyak2021introduction}, and characterize its variance reduction.
\item We describe an \textbf{efficient implementation} based on operator overloading and vectorization that enables dimensional peeking for simulations in C++ with minimal user effort. % TODO: add link later: \footnote{Available at \url{https://github.com/TODO}}
\item We report \textbf{results from extensive experiments} to evaluate the variance reduction, execution time overhead, and optimization performance of dimensional peeking in three discrete SBO problems against three popular meta-heuristics.
\end{itemize}

The remainder of the paper is structured as follows.
In Section~\ref{sec:background_and_related_work}, we provide background on gradient estimation techniques for discrete simulations and differentiate our method from existing variance reduction techniques.
Section~\ref{sec:dimensional_peeking} introduces the method of dimensional peeking and its efficient implementation.
In Section~\ref{sec:experiments}, we present experimental results to explore the method's variance reduction, computational overhead, and optimization progress over time in three SBO problems.
Section~\ref{sec:conclusions} provides a discussion of our results and future work and concludes the paper.

\section{Background and Related Work}
\label{sec:background_and_related_work}

In the following, we discuss how our approach relates to existing work on gradient estimation for discrete simulations and briefly cover established variance reduction techniques applicable in this context.

\subsection{Gradient Estimation for Discrete Simulations}

Simulations often involve jump discontinuities originating from conditional branching.
Although the output of a stochastic simulation involving branches may still be continuous in expectation, each simulation evaluation may observe a different control flow path.
The challenge in estimating gradients over such simulations lies in correctly accounting for the effects of the discontinuities based on a finite sample of trajectories.

\subsubsection{Traditional approaches}
Classical gradient estimators from the SBO literature rely on manual analysis and problem knowledge to derive unbiased gradient estimators.
In \emph{smoothed perturbation analysis}~\cite{fu2006gradient}, the objective function is separated into continuous parts through a problem-specific conditioning on suitable variables.
The \emph{likelihood ratio estimator}, also known as REINFORCE or score function estimator, applies the differentiation rule of the logarithm to allow unbiased gradient estimations with respect to parameters of known distributions~\cite{williams1992simple}.

\subsubsection{Automatic Differentiation}
Recent methods frequently rely on automatic differentiation (AD)~\cite{margossian2019review}, which propagates derivative information along the operations involved in a program by repeatedly applying the chain rule.
In forward-mode AD, an original program's variables are extended to carry a tangent with respect to an input variable in addition to the original variable value.
Reverse-mode AD, a special case of which is the well-known backpropagation algorithm~\cite{rumelhart1986learning}, records the operations and computes derivatives back-to-front once the forward execution has terminated.
As AD alone cannot account for discontinuities, a variety of works substitute jumps with smooth approximations such as logistic or sigmoid functions~\cite{newbury2024review,andelfinger2023towards,chopra2023agenttorch}.
A downside of these approaches is the difficulty of predicting and controlling the bias introduced by this form of smoothing~\cite{zhong2022differentiable}, which lacks a clear interpretation.

Some recent works propose methods to compute gradients without the need for manual derivations or problem-specific smoothing.
Arya et al.~proposed a form of forward-mode AD that computes unbiased gradients for stochastic programs that draw from a set of discrete distributions~\cite{arya2022automatic}.
Based on a custom chain rule, it suffices to propagate the variable values for a ``primal'' trajectory and a single alternative trajectory throughout the program.
An approach supporting arbitrary conditional branching was proposed by Kreikemeyer et al.~\cite{kreikemeyer2023smoothing}.
Their estimator records branch condition variables and their derivatives and combines them with pathwise AD gradients to account for the effects of branches.
As the methods relies on density estimations, the approach involves a variance-bias tradeoff.

While dimensional peeking does not rely on AD, its realization via operator overloading bears a loose similarity to forward-AD implementations via dual numbers~\cite{margossian2019review}, albeit with no similarities in the computation rules or interpretation of the output vectors.
The implementation shares the use of vectorization with our prior work on accelerating gradient estimation for agent-based simulations~\cite{andelfinger2025slight}.
However, aiming for speedup rather than variance reduction, this existing work focused on the efficient evaluation of sets of random parameter combinations, in contrast to the change in sampling granularity achieved by dimensional peeking.

\subsubsection{Stochastic Estimators}

Finally, stochastic black-box estimators employ random perturbations to compute smoothed gradient estimates from finite differences across the simulation output.
Simultaneous perturbation stochastic approximation~\cite{spall2002multivariate} is a gradient descent scheme that relies on central finite difference estimates using random perturbations, scaled by a decreasing series over the iterations.
A similar estimator described by Polyak~\cite{polyak2021introduction} employs forward differences and Gaussian perturbations and has been analyzed in the context of random search strategies by Nesterov et al.~\cite{nesterov2017random}.
We refer to this estimator as Polyak's Gradient Oracle (PGO).
Due to the external perturbations, the gradients returned by these stochastic estimators are unbiased with respect to a \emph{smooth approximation} of the original objective function.
For instance, Polyak's estimator reflects the gradients of the original objective function after convolution with a Gaussian kernel.
While the perturbations introduce a bias, these estimators are generically applicable without the need for prior analysis of the objective function or code adaptations as typically required for AD.
Stochastic estimators have been applied to convex and non-convex, smooth and non-smooth, as well as real, discrete, and mixed objectives~\cite{ghadimi2013stochastic,fang2018spider,ji2019improved,balasubramanian2022zeroth,zhang2023gradient,andelfinger2024automatic,wang2025simultaneous}.

In the present paper, we specialize Polyak's estimator for simulations over discrete decision variables and derive dimensional peeking based on this formulation.

\subsection{Variance Reduction Methods}

We briefly discuss how existing methods for variance reduction relate to dimensional peeking, restricting our discussion to works specific to stochastic black-box estimators.
This excludes generic approaches such as control variates, antithetic variates, Rao-Black\-well\-ization, and importance sampling~\cite{robert1999monte}, as these are orthogonal to and may be combined with our method.

Petersen et al.~\cite{petersen2024generalizing} evaluated different sampling schemes for the random perturbations used by stochastic gradient estimators.
Choosing perturbations based on low-discrepancy sequences rather than i.i.d.~sampling achieves better coverage of the neighborhood around the current solution.
Similarly, the coverage of the multi-dimensional input space can be improved by choosing orthogonal directions across dimensions~\cite{kozak2023zeroth,feng2023stochastic}.
By choosing dependent perturbations, these approaches sacrifice unbiasedness with respect to the smoothed gradient in favor of reduced variance.
In contrast, dimensional peeking covers the non-negligible portion of the perturbations' support on all input dimensions entirely, lifting the sampling from individual values of the discrete variables to classes of values that lead to the same control flow.

\section{Dimensional Peeking}
\label{sec:dimensional_peeking}

In the following, we derive the approach of dimensional peeking from an existing stochastic gradient estimator and characterize the achieved variance reduction.
Subsequently, we describe an efficient implementation of the required adaptations to lift the evaluation of a program with discrete inputs from the scalar level to classes of values that lead to the same control flow path.

\subsection{Derivation}
\label{sec:dimensional_peeking:subsec:derivation}

We start with the stochastic forward-differences gradient oracle described by Polyak~\cite{polyak2021introduction} and analyzed by Nesterov et al.~\cite{nesterov2017random}, which we refer to as Polyak's Gradient-Free Oracle (PGO).
Here, we adapt PGO for the discrete case of a function $f: \mathbb{Z}^d \rightarrow \mathbb{R}$ by drawing vectors $R$ of random perturbations from a discretized normal distribution:
$$
R \sim \text{Discrete-}\mathcal{N}(0, \sigma^2 I_d)
$$
with
$$
P(R \! = \! r) = \int_{r-0.5}^{r+0.5} \frac{1}{\sqrt{2\pi}\,\sigma} \, e^{-\frac{t^2}{2\sigma^2}} \, dt.
$$
In this setting, the PGO estimator is:
$$
g_{\text{PGO}} = \bigl(f(x + R) - f(x)\bigr)R\sigma^{-2}.
$$
PGO's expectation is the gradient of a discrete Gaussian approximation of $f$:
$$
\mathbb{E}[g_{\text{PGO}}] = \nabla_x \!\!  \sum_{r \in \mathbb{Z}^d} \!\! f(x + r)\, P(R \! = \! r)
$$
We will now derive a variant of PGO with the same expectation, but per-dimension variance that is typically lower, and at most equal.
For this, we express PGO's expectation over a single perturbation dimension $i$ for a given realization of all other dimensions.
For brevity, we express this conditioning using the function
$$f_{r_{-i}}(R_i) := f(r_1,\dots,r_{i-1},\,R_i,\,r_{i+1},\dots,r_d).$$
The expectation becomes:
$$
\mathbb{E}[g_{\text{PGO},i}] = \mathbb{E}[(f_{-i}(x_i + R_i) - f(x))\sigma^{-2}R_i].
$$
By the law of total expectation, equality of this per-variable expectation implies equality of the overall expectation.
We now partition the integers into equivalence classes $[y] \subseteq \mathbb{Z}$ according to an equivalence relation $\sim$ to group the perturbations on dimension $i$.
By the definition of expectation and since the equivalence classes are disjoint, we can write:
\begin{align*}
\mathbb{E}[g_{\text{PGO},i}] = \sum_{r_i \in \mathbb{Z}}{ \!\! P(R_i \! = \! r_i)\bigl(f_{-i}(x_i + r_i) - f(x)\bigr)\sigma^{-2}r_i} \\
= \sum_{[r_i] \in \mathbb{Z}/{\sim}}\,\sum_{r_i' \in [r_i]}{ \!\! P(R_i \! = \! r_i')\bigl(f_{-i}(x_i + r_i') - f(x)\bigr)\sigma^{-2}r_i'} \\
= \sum_{[r_i] \in \mathbb{Z}/{\sim}}{ \!\! P(R_i \in [r_i])}\mathbb{E}\left[{\bigl(f_{-i}(x_i + R_i) - f(x)\bigr)\sigma^{-2}R_i} | R_i \in [r_i]\right].
\end{align*}
The corresponding partial derivative oracle for dimension $i$ is
$$
g_{\text{PGO-DP},i} = \frac{1}{P(R'_i \! \in \! [R_i])} \!\! \sum_{r_i \in [R_i]}{ \!\!\!\! P(R'_i \! = \! r_i)\bigl(f_{-i}(x_i + r_i) \! - \! f(x)\bigr)\sigma^{-2}r_i},
$$
where $R'_i$ is a random variable independent of $R_i$ that follows the same distribution.

Applying this oracle to all independently perturbed dimensions yields a gradient oracle that considers all per-dimension perturbations in an equivalence class at once, weighted using the known perturbation probabilities.

In practice, we restrict the coverage of alternative perturbations by a radius $c$ around the primal perturbation for efficiency.
If a primal perturbation falls outside the radius, we fall back to the original PGO estimator for the current input dimension.
With appropriate scaling, limiting the coverage to parts of the encountered equivalence classes does not introduce any bias.
Let \( S([r_i]) \subseteq [r_i] \) denote the covered subset of the equivalence class \([r_i]\) and let
\[
p_S = P(R'_i \in S([r_i]) \mid R'_i \in [r_i]) 
     = \frac{P(R'_i \in S([r_i]))}{P(R'_i \in [r_i])}.
\]
Scaling by $p_S^{-1}$ preserves the expectation:
\begin{align*}
\mathbb{E}\Bigl[
  p_S^{-1}\!\!\!\!\!\!\sum_{r_i \in S([r_i])}
  \!\! P(R'_i = r_i)
  \bigl(f_{-i}(x_i + r_i) - f(x)\bigr)\sigma^{-2}r_i
  \,|\, R_i \in [r_i]
\Bigr]
=\\
\mathbb{E}\Bigl[\!\!\!\sum_{r_i \in [r_i]}
  \!\! P(R'_i = r_i)
  \bigl(f_{-i}(x_i + r_i) - f(x)\bigr)\sigma^{-2}r_i
  \,|\, R_i \in [r_i]
\Bigr].
\end{align*}
Hence, the modified estimator $g_{\text{PGO-DP},S,i} =$
\begin{align*}
  \frac{1}{P(R'_i \in S([R_i]))}
  \!\sum_{r_i \in S([R_i])}\!\!\!\!\!
  P(R'_i = r_i)
  \bigl(f_{-i}(x_i + r_i) - f(x)\bigr)\sigma^{-2}r_i
\end{align*}
has the same expectation as \( g_{\text{PGO-DP},i} \).

\subsection{Variance Reduction}
\label{sec:dimensional_peeking:subsec:variance_reduction}

We now consider the variance reduction achieved by PGO-DP under full coverage of an equivalence class.
Using the law of total variance, we can express the original PGO's variance on dimension $i$ as
\begin{align*}
\mathrm{Var}(g_{\text{PGO},i}) = \mathbb{E}\bigl[\mathrm{Var}\bigl(\bigl(&f_{-i}(x_i + R_i) - f(x)\bigr)R_i\sigma^{-2} | [R_i] \bigr)\bigr] + \\
\mathrm{Var}(\mathbb{E}\bigr[\bigl(&f_{-i}(x_i + R_i) - f(x)\bigr)R_i\sigma^{-2} | [R_i]\bigr]).
\end{align*}
The first summand is the expected variance \emph{within} an equivalence class, the second summand is the variance of the expectation \emph{across} equivalence classes.
As PGO-DP yields the same value for any perturbation within the same equivalence class, its variance is only
$$
\mathrm{Var}(g_{\text{PGO-DP},i}) = \mathrm{Var}(\mathbb{E}[\bigl(f_{-i}(x_i + R_i) - f(x)\bigr)R_i\sigma^{-2} | [R_i]]).
$$
The per-dimension variance reduction ratio is thus
$$
\frac{\mathrm{Var}(g_{\text{PGO},i})}{\mathrm{Var}(g_{\text{PGO-DP},i})} = 1 + \frac{\mathbb{E}[\mathrm{Var}(\bigl(f_{-i}(x_i + R_i) - f(x)\bigr)R_i\sigma^{-2} | [R_i] )]}{\mathrm{Var}(\mathbb{E}[\bigl(f_{-i}(x_i + R_i) - f(x)\bigr)R_i\sigma^{-2} | [R_i]])}.
$$
Note that this holds for any realization of the independent perturbations on other dimensions.
If $f$ involves additional random variables beyond $R$, as is the case in many simulations, the above argumentation applies by the substitutions $f(x) \mapsto f(x | \omega_0)$ and $f_{r_{-i}}(R_i) \mapsto f_{r_{-i}}(R_i | \omega_1)$, where $\omega_0, \omega_1 \in \Omega$ are tuples of realizations of the random variables and $\Omega$ their support.

An important case is $\mathbb{Z}/{\sim} = \{\mathbb{Z}\}$, i.e., all perturbations on a dimension fall into the same class.
Then, the variance across classes and thus PGO-DP's total variance stemming from the perturbations is $0$, whereas PGO retains the variance within the class.

The impact of the coverage radius $c$ on the variance reduction is assessed in Section~\ref{sec:experiments:subsec:impact_of_coverage_radius}.

\subsection{Control Flow Equivalence}
\label{sec:dimensional_peeking:subsec:control_flow_equivalence}

The above derivation generically applies to any equivalence relation on the perturbations.
However, attaining a variance reduction requires us to determine the function outputs for multiple perturbations in a class.
We achieve this without explicit additional sampling by defining the equivalence relation to represent equivalence in terms of control flow.
More formally, let $\text{Path}(x) = \langle n_1, n_2, \dots, n_k \rangle$ be the sequence of nodes in the given program's control-flow graph followed for an input vector x.
Input-dependent loops are represented by repeated occurrences of the corresponding nodes in the path sequence.
Similarly to $f_{-i}$ above, we define
$$
\text{Path}_{-i}(r'_i) = \text{Path}\left(x+(r_1,\dots,r_{i-1},r_i',r_{i+1},\dots,r_d)\right)
$$
This is the path followed given the primal perturbed input vector on all dimensions $j \ne i$ and the alternative perturbed input $x_i + r'_i$ on dimension $i$.
We define two perturbations $r_i$, $r'_i$ to be \emph{control flow-equivalent} if $\text{Path}_{-i}(r_i) = \text{Path}_{-i}(r'_i)$
For clarity, we note that control-flow equivalence does not imply equivalence in output, as illustrated by the following simple program snippet.
\begin{verbatim}
  if x[0] > 0:
    return x[0]
  ...
\end{verbatim}
Here, although any input \texttt{x[0]} $ > 0$ follows the same control flow path, the output still varies with \texttt{x[0]}.

\subsection{Implementation}
\label{sec:dimensional_peeking:subsec:implementation}

To form a practicable gradient estimator from our definitions of PGO-DP and control flow equivalence, two requirements must be met:
First, each program evaluation must be extended beyond the current primal perturbed input vector $x + r$ to alternative perturbations along each input dimension.
Second, for each alternative perturbation $r'_i$ along dimension $i$, control-flow equivalence with $r_i$ must be determined.
An efficient implementation of these mechanisms is essential to ensure that a faster convergence through reduced variance is not offset by the function evaluation overhead.

Although we describe our implementation using C++ constructs, the same ideas apply to other languages that support vectorization and operator overloading, such as Python and Julia.

\subsubsection{Perturbed Arithmetic}

We extend the execution of a program to several perturbations of the decision variables by altering the program to be evaluated on a \emph{vector} per decision variable reflecting several perturbations.
For instance, let the original three-dimensional input vector be $x = [3\ 1\ 5]$ and the primal perturbed input vector $x + r = [2\ 1\ 7]$.
The range of considered alternative perturbations around each dimension of $x$ is determined by the coverage radius $c$.
For $c = 2$, the perturbed input vector is translated to the three vectors $[ 1\ \scalebox{1.1}{\textbf{2}}\ 3\ 4\  5 ], [ -1\  0\  \scalebox{1.1}{\textbf{1}}\ 2\  3 ], [ 3\  4\  5\  6\  \scalebox{1.1}{\textbf{7}} ]$, bold numbers indicating the primal perturbed values.
We now propagate the effects of the perturbations through the arithmetic operations of the program, during which new vectors are calculated that may depend on perturbed decision variables on one or more dimensions.
For notational consistency, we denote the dependence on dimension $i$ using a subscript $x[i - 1]$, e.g., $[ 1\  \scalebox{1.1}{\textbf{2}}\ 3\ 4\  5 ]_{x[0]}$ for the first input dimension.

For unary operations such as absolute value, negation, and exponentiation, the arithmetic over perturbed variables trivially translates to element-wise arithmetic over all input dependencies and all perturbed values per dependency.
Similarly, binary operations between a perturbed variable and a constant, or between two perturbed variables with the same dependencies translates to element-wise operations for the pairs of scalars corresponding to the same dependency and perturbation.
For instance:
$$[ 1\  \scalebox{1.2}{\textbf{2}}\ 3\ 4\  5 ]_{x[0]} \times [ 3\ \scalebox{1.2}{\textbf{5}}\ 7\ 9\ 11 ]_{x[0]} = [ 3\  \scalebox{1.2}{\textbf{10}}\ 21\ 36\  55 ]_{x[0]}$$

The output of a binary operation between two perturbed variables with differing dependencies is a new perturbed variable carrying the \emph{union} of the decision variables' dependencies.
Dependencies on the same decision variable are handled in an element-wise fashion as described above.
For dependencies present in only one perturbed operand $v_0$ but not in another $v_1$, the binary operation is computed as the element-wise operation between $v_0$ perturbed values and $v_1$'s scalar primal value.
The following is an example of a multiplication between perturbed variables:
\vspace*{-0.2cm}
\[
[ 1\  \scalebox{1.2}{$\mathbf{2\ }$} 3\  4\  5 ]_{x[0]} \times [ -1\  0\  \scalebox{1.2}{$\mathbf{1\ }$} 2\  3 ]_{x[1]} = 
\begin{array}{r@{\hskip 0pt} r@{\hskip 2.5pt} r@{\hskip 2.5pt} r@{\hskip 2.5pt} r@{\hskip 2.5pt} r@{\hskip 2.5pt} r@{\hskip 0pt} r}
\bigl[ & 1 & \scalebox{1.2}{$\mathbf{2}$} & 3 & 4 & 5 & \bigr]_{x[0]} \\
\bigl[ & -2 & 0 & \scalebox{1.2}{\(\mathbf{2}\)} & 4 & 6 & \bigr]_{x[1]}\\
\end{array}
\vspace{-0.05cm}
\]

We implement the perturbed arithmetic by introducing a new data type \texttt{pfloat} (perturbed floating point number) that carries the scalar primal variable value together with the perturbed values.
This approach bears superficial similarities to forward-mode AD, where variables are extended to include their tangents with respect to the program inputs.
However, instead of tangents, dimensional peeking propagates alternative intermediate variable values under different input perturbations.

The \texttt{pfloat} type implements arithmetic operations and comparisons via operator overloading, allowing it to be used in the same manner as C++'s primitive floating point types.
The handling of dependencies on $d$ decision variables with $n$ perturbations each could be supported by carrying out element-wise arithmetic over an $d \times n$ matrix.
However, in the common case of variables that depend on only a few decision variables, this dense representation would incur many redundant computations.
We choose a sparse representation in which each \texttt{pfloat} carries a dynamically growing array holding one vector of perturbed values per dependency, and a single vector of pointers representing the dependencies.
To avoid the cost of dynamic memory allocation, we employ a stack-based dynamic array type.
While its interface is that of a C++ standard template library \texttt{vector}, the array reserves space for the maximum of $d$ vectors on the stack and handles item insertion purely by incrementing a size variable and item assignment.

In our sparse representation, the output from a binary operations depends on the union of the decision variables' dependencies.
We first search for $a$'s dependencies in $b$ to handle the intersection of the dependencies by an element-wise vector operation.
Dependencies not found in $b$ are handled as operations between $a$'s perturbation vector and $b$'s primal value.
In doing so, we mark the found intersecting dependencies in $b$'s dependency array.
The unmarked dependencies are present only in $b$ and are handled as operations between $a$' primal value and $b$'s perturbation vector.

As the dependency arrays are unordered, the time complexity of this process is in $O(mn)$ time, $m$ and $n$ being the lengths of $a$ and $b$'s dependency arrays.
Several simple heuristics are applied to maintain low cost:
\begin{itemize}
\item If one of the operator has no dependencies, the operation is handled as an operation between a \texttt{pfloat} and a scalar.
\item If $b$ has more dependencies than $a$, we swap the search order to reduce the cost of checking for unmarked operations.
\item To accelerate operations between operands with similar dependencies, we check for dependencies at identical indexes first before resorting to linear search.
\end{itemize}

In practice, the cost depends not only on the number of dependencies and perturbations, but also on the proportion of operations on \texttt{pfloat} instances in the program.
The overhead for three simulation models from the literature is evaluated in Section~\ref{sec:experiments}.

\subsubsection{Determining Control Flow-Equivalence}

\begin{algorithm}[b]
\caption{Overloaded comparison operator on \texttt{pfloat}.}
\label{alg:comparison}
\begin{algorithmic}
\small
\Function{cmp}{pfloat a, float b}
    \State t\_primal $\gets \text{cmp}(\text{a.primal}, \text{b})$\hspace{0.1cm}\textit{// scalar-scalar comparison}
    \For{$d \gets 0$ \textbf{to} $|\text{a.value\_vecs}|-1$}\hspace{0.1cm}\textit{// iterate dependencies}
        \State \textit{// vector-scalar comparison:}
        \State t\_perturbations $\gets (\text{cmp}(\text{a.value\_vecs}[d], \text{b}) == \text{t\_primal})$
        \State \vspace{-0.2cm}
        \State \textit{// update control-flow equivalences:}
        \State $\text{a.equivalence\_vecs}[d] \gets \text{a.equivalence\_vecs[d]} \ \&$
        \State \hspace{3.0cm}t\_perturbations
    \EndFor
    \State \Return t\_primal
\EndFunction
\end{algorithmic}
\end{algorithm}

In a deterministic imperative program (and, equivalently, in a stochastic program with fixed realizations of all random variables) the observed path depends solely on input-dependent control flow in the form of parameter-dependent branching, looping, and jumps.
The parameter dependence manifests in conditional expressions that are functions of input parameters.
Hence, to establish control flow equivalence between $r'_i$ and $r_i$ throughout a program execution based on $r_i$, it suffices to ensure that all encountered conditional expressions evaluate to the same truth value for $r'_i$ as for $r_i$.

As described above, the \texttt{pfloat} type contains a dynamic array of pointers to its dependencies.
More specifically, each element of the array points to a vector of boolean values representing control flow-equivalence with the primal perturbation for a decision variable.
Before evaluating the program, all booleans are initialized to \texttt{true}.
On each comparison involving a \texttt{pfloat}, control flow-equivalence is determined for each dependency, and the booleans are updated accordingly.
A comparison operator $\text{cmp}(a, b)$, where $a$ is a \texttt{pfloat} and $b$ a primitive numerical type, is overloaded according to pseudo code shown in Algorithm~\ref{alg:comparison}.

Here, the truth value of the comparison for the primal value is delegated to an ordinary scalar-scalar comparison.
Subsequently, vectors of truth values are gathered by vectorized comparisons for the perturbation vectors associated with each dependency.
After each vectorized comparison, the boolean vector of control flow-equivalences is updated.
The program execution continues for all perturbations, but finally, only the perturbations identified as control flow-equivalent with the primal perturbation contribute to the gradient estimate, scaled by the perturbation probabilities.

\begin{figure}[t]
  \centering
  \includegraphics[width=0.46\textwidth]{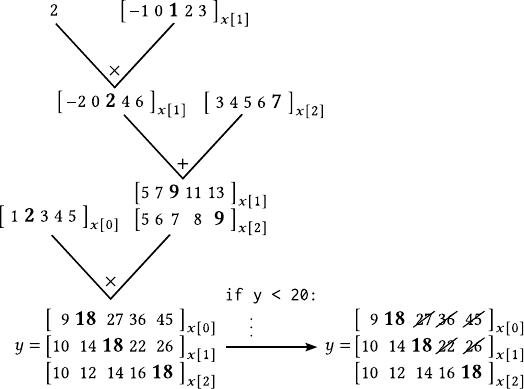}
  \caption{Dimensional peeking for a program comprised of arithmetic and a conditional branch, evaluated at the primal perturbed input $x = [2\  1\  7]$. Numbers in brackets represent the variable values under different perturbations, with the primal value shown in bold. Subscripts indicate the perturbed decision variables the value depends on. Evaluating the conditional statement \textnormal{\texttt{if y < 20}} rules out values on two of the three dimensions (crossed-out numbers).}
  \label{fig:computation_graph}
\end{figure}

To illustrate an overall program execution under dimensional peeking, we consider the following program snippet:
\begin{verbatim}
  y = x[0] * (2 * x[1] + x[2])
  if y < 20:
      ...
\end{verbatim}
Figure~\ref{fig:computation_graph} shows the vectorized arithmetic and identification of control flow-equivalence for the input vector $x = [2\  1\  7]$.

\section{Experiments}
\label{sec:experiments}

Our experiments study the variance reduction achieved using dimensional peeking, its execution time overhead, and finally its benefits when solving three optimization problems in comparison to gradient descent via the original PGO estimator and three popular meta-heuristics.

All execution time measurements employ the same codebase for PGO and PGO-DP, relying on C++ templating to switch between native floating point numbers and our \texttt{pfloat} type.
The perturbed arithmetic is carried out in a vectorized manner via single-instruction, multiple-data operations implemented via the Fastor library~\cite{poya2017high}.
The baselines in the optimization experiments include the genetic algorithm (GA) implementation of the pyeasyga module\footnote{\url{https://github.com/remiomosowon/pyeasyga}}, particle swarm optimization (PSO) via the PySwarms toolkit~\cite{miranda2018pyswarms}, and covariance matrix adaptation evolution strategy (CMA-ES)~\cite{auger2012tutorial} via the pycma module\footnote{\url{https://github.com/CMA-ES}}.
All baselines rely on native floating point numbers.

The experiments were carried out on a machine equipped with an AMD EPYC 7742 CPU and 256GiB RAM running Ubuntu 20.04.6.

\subsection{Optimization Problems}
\label{sec:experiments:subsec:optimization_problems}

We consider SBO problems using three models from the literature:

\textbf{\textsc{CityFlow}} is a microscopic traffic simulator targeted towards large-scale scenarios and traffic control applications~\cite{zhang2019cityflow}. The simulation engine is written in C++, allowing us to apply dimensional peeking by introducing our \texttt{pfloat} data type. We consider a traffic control problem over a scenario covering 4 $\times$ 4 intersections in Hangzhou, China\footnote{\url{https://traffic-signal-control.github.io/}}. The objective is the average traveled distance. The simulation involves 144 decision variables representing traffic light phase durations in seconds. Stochasticity is introduced by applying uniform random offsets in $\{0, \ldots, 10\}$s to the trip start times.

\textbf{\textsc{Hotel}} is a revenue maximization problem from the SimOpt suite of SBO problems~\cite{eckman2023simopt}. The goal is to maximize a hotel's revenue by adjusting the numbers of available ``products'' comprised of sequences of days in a week and two possible fares. The revenue is counted for one week after one week of warmup. Customers arrive randomly according to product-specific Poisson processes. As the products overlap in time, each successful booking reduces other products' availability. The problem involves 56 decision variables.

\textbf{\textsc{DynamNews}} is also part of the SimOpt suite and represents a newsvendor problem with dynamic demand. Each arriving customer assigns a score to each available product as a sum of a constant product-specific utility and a Gumbel random variable. Each customer chooses the product with the highest utility among the products still in stock. The objective is the overall revenue given by the sum of the sold products' prices, subtracting their cost. The decision variables are the products' prices and the initial inventory levels. We configure the constants according to a setting from the SimOpt web site\footnote{\url{https://simopt.readthedocs.io/en/development/models/dynamnews.html}}, extending it to 3\,000 customers and 1\,000 products, corresponding to 1\,000 decision variables.

A key difference among the problems is that in \textsc{CityFlow} and \textsc{Hotel}, the objective function value depends purely on control flow, i.e., if two sets of decision variable values follow the same control flow, the output is identical.
Hence, as discussed in Section~\ref{sec:dimensional_peeking:subsec:derivation}, PGO-DP reduces the variance to that across classes of control-flow equivalent decision variable values.
In contrast, perturbations to the products' prices in \textsc{DynamNews} directly propagate to the objective value, even if the control flow is unaffected.

The reported measurements regarding correctness, variance reduction and gradient estimation times are averages over $10^5$ repetitions for \text{CityFlow} and \textsc{DynamNews}, and $10^6$ for \textsc{Hotel}.
Optimization curves show averages across 30 replications starting from random parameter combinations for each combination of model and optimizer hyperparameters.

\subsection{Verification}
\label{sec:experiments:subsec:verification}

We verified the correctness and numerical stability of our implementation by comparing derivative estimates between PGO and PGO-DP.
Table~\ref{tbl:unbiasedness} shows the mean difference between the estimates with $\sigma = 1$ across all input dimensions with 99\% confidence intervals.
Setting a coverage radius of $c = 15\sigma$ reflects full coverage of the perturbation probabilities representable as a single precision floating point number.
In line with the theoretical results from Section~\ref{sec:dimensional_peeking:subsec:derivation} showing unbiasedness with respect to the smoothed objective, the statistical results do not indicate a deviation of the estimators' expectations, independently of the coverage radius.

\begin{table}[t]
\caption{Verification of our implementation. The percentage difference between PGO and PGO-DP does not indicate a bias at the 1\% significance level for any coverage radius $c$.\vspace{-0.3cm}}
\small
\begin{tabular*}{0.47\textwidth}{@{\extracolsep{\fill}}c r@{ $\pm$\hspace{-0.35cm}}l r@{ $\pm$\hspace{-0.35cm}}l r@{ $\pm$\hspace{-0.35cm}}l}
\toprule
c & \multicolumn{2}{c}{\textsc{CityFlow}} & \multicolumn{2}{c}{\textsc{Hotel}} & \multicolumn{2}{c}{\textsc{DynamNews}} \\
\midrule
{1$\sigma$} & -0.023 & 0.217 & 0.014 & 0.107 & -0.044 & 0.101\\
{3$\sigma$} & 0.133 & 0.199 & -0.011 & 0.090 & 0.012 & 0.096\\
{5$\sigma$} & 0.011 & 0.197 & 0.016 & 0.090 & 0.001 & 0.096\\
$\mathbf{15\boldsymbol{\sigma}}$ & $\mathbf{-0.082}$ & $\mathbf{0.196}$ & $\mathbf{-0.021}$ & $\mathbf{0.090}$ & $\mathbf{0.006}$ & $\mathbf{0.095}$\\
\bottomrule
\end{tabular*}
\label{tbl:unbiasedness}
\vspace{-0.1cm}
\end{table}

\begin{table}[t]
\caption{Analytical and measured variance reduction rate (VRR) between PGO and PGO-DP for the Heaviside function. Larger smoothing factors $\sigma$ yield a larger VRR.\vspace{-0.3cm}}
\small
\begin{tabular*}{0.47\textwidth}{@{\extracolsep{\fill}}c c c c c}
\toprule
         & Expected          & Var. across  & Analytical & Measured \\
$\sigma$ & in-class var. &  class means &  VRR &          VRR \\
\midrule
1 & 0.069 & 0.327 & 1.212 & 1.203\\
2 & 0.122 & 0.232 & 1.525 & 1.524\\
4 & 0.151 & 0.193 & 1.781 & 1.799\\
8 & 0.166 & 0.176 & 1.946 & 1.939\\
\bottomrule
\end{tabular*}
\label{tbl:vrr_heaviside}
\vspace{-0.2cm}
\end{table}

To verify that our implementation of PGO-DP achieves the variance reduction characterized analytically in Section~\ref{sec:dimensional_peeking:subsec:variance_reduction}, we estimate smoothed derivatives of the Heaviside step function
$$
H(x) = 
\begin{cases}
0, & x < 0 \\
1, & x \ge 0
\end{cases}.
$$
As stated in Section~\ref{sec:dimensional_peeking:subsec:variance_reduction}, PGO's variance can be expressed as the sum of the expected variance \emph{within} a class of control flow-equivalent values ($x + R < 0$ or $x + R \ge 0$) and the variance of the expectation \emph{across} classes.
With PGO-DP and sufficiently large coverage radius $c$, the in-class variance is eliminated.
We analytically determined the two summands for $x = 0$ with $c = 15\sigma$ and different smoothing factors $\sigma$ based on the known perturbation probabilities (cf.~Appendix~\ref{app:variance_reduction_for_heaviside_step_function}) and compared the results to measurements via PGO and PGO-DP over $10^5$ estimations.
The results in Table~\ref{tbl:vrr_heaviside} show that the measured variance reduction ratio (VRR) is in line with the analytical results.
With larger smoothing factor $\sigma$, the in-class variance increases, whereas the variance across class means decreases, yielding a larger VRR.

\subsection{Impact of Coverage Radius}
\label{sec:experiments:subsec:impact_of_coverage_radius}

In contrast to the estimates' expectation, the variance reduction achieved by PGO-DP is affected by the coverage radius $c$.
Table~\ref{tbl:vrr_simulation_models} shows the VRR for the three simulation models used in our optimization experiments using $\sigma = 1$.
The variance reduction is substantial for all models, the largest VRR being $7.9$ for the \textsc{Hotel} model.
Beyond $c = 3\sigma$, only modest increases in VRR are observed, making this value sufficient for the optimization experiments of Section~\ref{sec:experiments:subsec:optimization_progress}.

\begin{table}[t]
\small
\caption{Variance reduction ratios for the simulation models, varying the coverage radius $c$.\vspace{-0.3cm}}
\begin{tabular*}{0.47\textwidth}{@{\extracolsep{\fill}}c c c c}
\toprule
c & \textsc{CityFlow} & \textsc{Hotel} & \textsc{DynamNews} \\
\midrule
${1\sigma}$ & 1.36 & 1.66 & 1.17\\
$\mathbf{3\boldsymbol{\sigma}}$ & \textbf{2.55} & \textbf{7.53} & \textbf{1.45}\\
${5\sigma}$ & 2.62 & 7.88 & 1.46\\
$\mathbf{15\boldsymbol{\sigma}}$ & \textbf{2.62} & \textbf{7.90} & \textbf{1.44}\\
\bottomrule
\end{tabular*}
\label{tbl:vrr_simulation_models}
\end{table}

The coverage radius also affects PGO-DP's computational cost.
Table~\ref{tbl:slowdown} shows that since $c$ affects the size of the vectors involved in the perturbed arithmetic, the overhead increases moderately with larger $c$.
When extending the coverage to the full support of the perturbations up to machine precision by setting $c = 15\sigma$, the maximum overhead is 43\% for \textsc{Hotel}.
With $c = 3\sigma$, the maximum overhead is 28\%.
The comparatively larger overhead for \textsc{Hotel} and \textsc{DynamNews} is explained by their larger proportion of operations on \texttt{pfloat} variables.

We also measured the overhead in memory consumption, which depends on the number of resident \texttt{pfloat} values during a simulation run and the size of each \texttt{pfloat} according to the coverage radius.
For \textsc{CityFlow}, memory consumption increased by factors of 2.2 to 2.6 with $c = 1\sigma$ and $c = 15\sigma$ compared to PGO.
In contrast, \textsc{Hotel} and \textsc{DynamNews} showed increases of 2\% or less.

\begin{table}
\caption{Slowdown factors, varying the coverage radius $c$.\vspace{-0.3cm}}
\small
\begin{tabular*}{0.47\textwidth}{@{\extracolsep{\fill}}c c c c}
\toprule
c & \textsc{CityFlow} & \textsc{Hotel} & \textsc{DynamNews} \\
\midrule
${1\sigma}$ & ${1.11 \pm 0.00}$ & ${1.26 \pm 0.00}$ & ${1.22 \pm 0.00}$\\
$\mathbf{3\boldsymbol{\sigma}}$ & $\mathbf{1.09 \pm 0.00}$ & $\mathbf{1.28 \pm 0.00}$ & $\mathbf{1.22 \pm 0.00}$\\
${5\sigma}$ & ${1.12 \pm 0.00}$ & ${1.23 \pm 0.00}$ & ${1.35 \pm 0.00}$\\
$\mathbf{15\boldsymbol{\sigma}}$ & $\mathbf{1.12 \pm 0.00}$ & $\mathbf{1.43 \pm 0.00}$ & $\mathbf{1.39 \pm 0.00}$\\
\bottomrule
\end{tabular*}
\label{tbl:slowdown}
\end{table}

\begin{figure}[t]
  \centering
  \begin{subfigure}[b]{0.49\textwidth}
    \includegraphics[width=\textwidth]{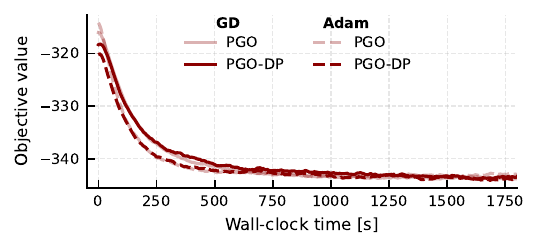}
    \vspace{-0.6cm}
    \caption{\textsc{CityFlow}}
  \end{subfigure}
  \begin{subfigure}[b]{0.49\textwidth}
    \includegraphics[width=\textwidth]{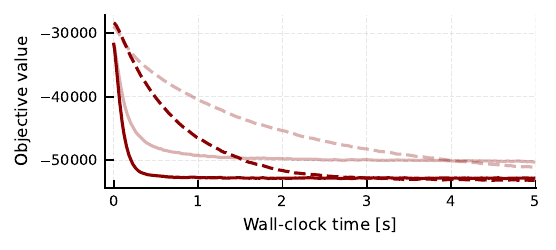}
    \vspace{-0.6cm}
    \caption{\textsc{Hotel}}
  \end{subfigure}
  \begin{subfigure}[b]{0.49\textwidth}
    \includegraphics[width=\textwidth]{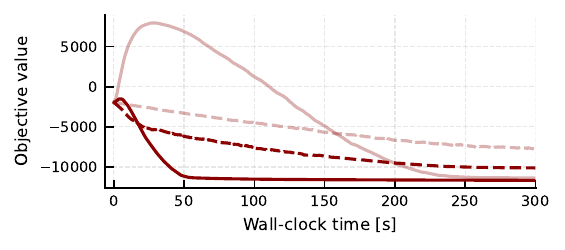}
    \vspace{-0.6cm}
    \caption{\textsc{DynamNews}}
  \end{subfigure}
  \caption{Comparison of the optimization progress over time between PGO and PGO-DP with $\sigma = 1$ and learning rates of $0.01$ and $0.1$ for SGD and Adam. For \textsc{Hotel} and \textsc{DynamNews}, PGO-DP significantly outperforms PGO.}
  \label{fig:optimization_cmp}
  \vspace{-0.2cm}
\end{figure}

\subsection{Optimization Progress}
\label{sec:experiments:subsec:optimization_progress}

To evaluate the effects of PGO-DP when solving overall optimization problems, we compare its convergence behavior against four baseline methods: gradient descent using PGO, genetic algorithm (GA), particle swarm optimization (PSO), and covariance matrix adaptation evolution strategy (CMA-ES).

To account for the optimizers' dependence on hyperparameters, we conducted a limited parameter sweep.
For GA, we configured a population size of 30 and enabled elitism.
The mutation function shifts a decision variable by a Gaussian random variable with $\mu = 0$ and $\sigma \in \{1, 2, 4\}$.
PSO uses 30 particles and the parameters $c_1 = 1.5, c_2 = 1.5, w \in \{ 0.4, 0.65, 0.9 \}$.
For CMA-ES, the standard deviation was set to 1, 2, and 4.
Similarly, for PGO and PGO-DP, we set the smoothing factor $\sigma$ to 1, 2, and 4.
The gradient estimates are applied in traditional gradient descent (GD) and using the Adam optimizer~\cite{adam2014method} with learning rates in $\{ 0.001, 0.05, 0.01 \}$ and $\{ 0.01, 0.05, 0.1 \}$.

As the execution time per function evaluation and the number of evaluations per optimization step differ between the methods, we plot the optimization progress over wall-clock time.
All problems are cast as minimization problems by negating the objective values.

We first compare PGO and PGO-DP using the same smoothing factor of $\sigma = 1$ and learning rates of 0.01 and 0.1 for GD and Adam.
Figure~\ref{fig:optimization_cmp} shows no substantial difference in convergence between PGO and PGO-DP in the \textsc{CityFlow} problem.
Although Table~\ref{tbl:vrr_simulation_models} showed a VRR of 2.6, the absolute variance of \textsc{CityFlow} is around three orders of magnitude lower than that of the other problems.
We conjecture that for this reason, the optimization progress is not sensitive to further variance reductions.

In contrast, both \textsc{Hotel} and \textsc{DynamNews} benefit strongly from the reduced variance, both with GD and Adam.
For \textsc{DynamNews}, we observe that initially, GD is vastly misled by PGO's higher-variance estimates, leading to a strong decline in solution quality.
With PGO-DP, a minuscule initial decline is observed, after which the solution quality improves quickly.
The solution quality at the end of the time budget is consistently higher with PGO-DP.

\begin{figure}[h!]
  \centering
  \begin{subfigure}[b]{0.49\textwidth}
    \includegraphics[width=\textwidth]{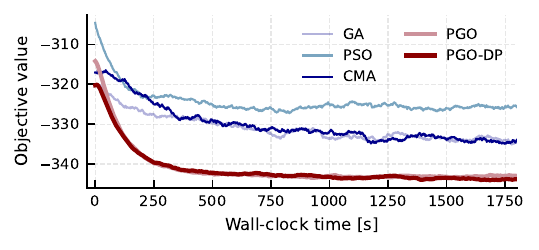}
    \vspace{-0.6cm}
    \caption{\textsc{CityFlow}}
  \end{subfigure}
  \begin{subfigure}[b]{0.49\textwidth}
    \includegraphics[width=\textwidth]{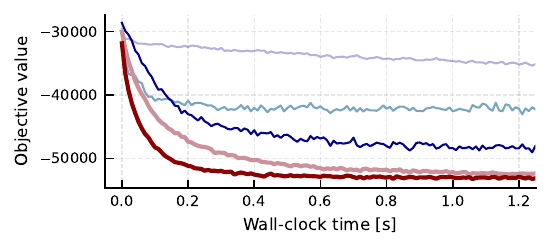}
    \vspace{-0.6cm}
    \caption{\textsc{Hotel}}
  \end{subfigure}
  \begin{subfigure}[b]{0.49\textwidth}
    \includegraphics[width=\textwidth]{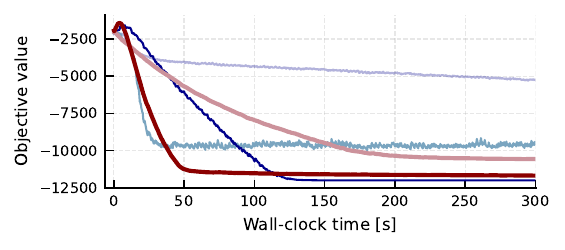}
    \vspace{-0.6cm}
    \caption{\textsc{DynamNews}}
  \end{subfigure}
  \caption{Optimization progress over time with the best hyperparametrization for each method. Gradient descent via PGO and PGO-DP excels at \textsc{Hotel} and is competitive at \textsc{DynamNews}. PGO-DP consistently outperforms PGO.}
  \label{fig:optimization_best}
\end{figure}

Finally, we plot the results with the best-performing hyperparameter combination for PGO-DP and all baselines in Figure~\ref{fig:optimization_best}.
We define the better optimization performance as achieving a lower area under the optimization curve.
For PGO and PGO-DP, we report results with the better-performing optimizer, GD or Adam.
In the \textsc{CityFlow} problem, PGO and PGO-DP outperform the meta-heuristic baselines, but comparing PGO and PGO-DP, no significant improvement is observed.

For \textsc{Hotel}, we constrain the time axis to more clearly show the initial optimization steps, during which most of the progress occurs.
Here, PGO-DP achieves the fastest convergence and reaches the best solution quality of all methods.
The objective value of -50\,000 is reached more than a factor 2 sooner than with PGO.
Both PGO-DP and PGO substantially outperform the meta-heuristics.

In \textsc{DynamNews}, PSO and PGO-DP achieved the fastest initial improvement, with PGO-DP converging to a much better solution quality after about 50s.
CMA-ES required about 125s to reach convergence, but achieved a slightly better solution quality.
As we saw in Figure~\ref{fig:optimization_cmp}, the lower variance of PGO-DP allows for fast convergence at higher learning rates compared to PGO.
Accordingly, the best-performing learning rate for PGO-DP was 0.001 using GD, whereas PGO performed best with a learning rate of 0.01.

\begin{table}[t]
\small
\caption{Speedup of PGO-DP over PGO in approaching the best found objective value. PGO never reached 99\% of PGO-DP's improvement over the starting solution.\vspace{-0.3cm}}
\centering
\begin{tabular*}{0.47\textwidth}{@{\extracolsep{\fill}}c c c c c c}\toprule
Simulation model & 75\% & 90\% & 95\% & 99\% \\
\midrule
\textsc{CityFlow} & 1.02 & 0.96 & 0.96 & - \\
\textsc{Hotel} & 2.60 & 2.77 & 3.09 & - \\
\textsc{DynamNews} & 4.10 & - & - & - \\
\bottomrule
\end{tabular*}
\label{tbl:percentage_of_best}
\end{table}

In Table~\ref{tbl:percentage_of_best}, we quantify the faster optimization progress with PGO-DP by comparing the times after which a certain improvement over the initial solution quality has been reached.
Using PGO-DP as the reference, the percentage improvement at optimization step $s$ is computed as $\frac{100y_s}{y_\Omega - y_A}$, where $y_s$, $y_A$, and $y_\Omega$ are the objective values at step s, the beginning, and the end of PGO-DP's optimization trajectory.
We compare the best-performing hyperparametrization of PGO and PGO-DP for each problem, including the choice of optimizer.
As previously observed in Figure~\ref{fig:optimization_cmp}, the results for \textsc{CityFlow} differ only slightly between PGO and PGO-DP.
For the other problems, the differences in solution qualities are more substantial, and PGO-DP reaches almost all improvement levels significantly faster, the largest speedup being 4.1 at 75\% improvement for \textsc{DynamNews}.
The baseline PGO was not able to reach 99\% of PGO-DP's improvement within the time budget for any of the models.

\section{Conclusions}
\label{sec:conclusions}

We presented dimensional peeking, a novel variance reduction method for zeroth-order discrete optimization via simulation.
Our experiments showed that at an execution time overhead between 9\% and 28\%, dimensional peeking reduces the variance of gradient estimates for three simulation-based optimization problems by factors between 1.5 to 7.5 over estimations on the scalar level.
In two of the three problems, the reduced variance translated to substantially improved convergence behavior.
Overall, the presented approach improves the competitiveness of zeroth-order optimization via gradient descent as compared to GA, PSO, and CMA-ES.

Our future work aims to extend the scope of dimensional peeking beyond gradient descent alone.
One promising avenue is its integration into optimization methods for non-convex problems.
For instance, trust-region algorithms such as TRON~\cite{lin1999newton} and ASTRO-DF~\cite{shashaani2018astro} could benefit from low-variance gradient estimates when constructing local approximations of the objective function.

Finally, outside of static parameter optimization, dimensional peeking could accelerate the training of reinforcement learning policies over discrete action spaces by enhancing or substituting established high-variance gradient estimator such as REINFORCE~\cite{williams1992simple}.

\begin{acks}
This research is supported by the National Research Foundation, Singapore under its AI Singapore Programme (AISG Award No: AISG3-RP-2022-031). 
\end{acks}

\bibliographystyle{ACM-Reference-Format}
\bibliography{references}

\appendix
\section{Variance Reduction for Heaviside Step Function}
\label{app:variance_reduction_for_heaviside_step_function}

In Section~\ref{sec:experiments:subsec:verification}, we verified our implementation by comparison to analytical determined variance reductions for the Heaviside step function $H$.
Here, we briefly show the steps to compute the references values for derivative estimates at $x = 0$.
Let $R \sim \text{Discrete-}\mathcal{N}(0,\sigma^2)$.
The probabilities for negative and positive perturbations are:
\[
p = \sum_{k=-\infty}^{-1} \frac{\exp(-k^2/(2\sigma^2))}{\sum_{m=-\infty}^{\infty} \exp(-m^2/(2\sigma^2))}, \qquad
\]

\[
q = 1 - p.
\]

As $H(x) - H(0) = 0 \quad \forall x \in \mathbb{N}^+$, positive perturbations do not contribute to the derivative estimates.
The mean and variance in the negative class are:

\[
\mu_{<0} = \frac{1}{p} \sum_{k=-\infty}^{-1} (-k) \frac{\exp(-k^2/(2\sigma^2))}{\sum_{m=-\infty}^{\infty} \exp(-m^2/(2\sigma^2))}, \quad
\]
\[
\sigma^2_{<0} = \frac{1}{p} \sum_{k=-\infty}^{-1} k^2 \frac{\exp(-k^2/(2\sigma^2))}{\sum_{m=-\infty}^{\infty} \exp(-m^2/(2\sigma^2))} - \mu_{<0}^2.
\]

Decomposing the total variance into the in-class variance and the variance of the expectation across classes, we obtain the variance reduction rate (VRR):

\[
\operatorname{Var}(\text{PGO}) = p \sigma^2_{<0} + p q \mu_{<0}^2, \qquad
\]
\[
\text{VRR} = 1 + \frac{p \sigma^2_{<0}}{p q \mu_{<0}^2}.
\]

\end{document}